\documentclass[11pt,a4paper,twocolumn,twoside]{article}
\usepackage[margin=1.5cm]{geometry}
\usepackage{cite}
\usepackage[pdftex]{graphicx}

\usepackage[utf8]{inputenc} % allow utf-8 input
\usepackage[T1]{fontenc}    % use 8-bit T1 fonts
\usepackage{lmodern}
\usepackage{hyperref}       % hyperlinks
\hypersetup{colorlinks,linkcolor=blue, citecolor=red}
\usepackage{url}            % simple URL typesetting
\usepackage{booktabs}       % professional-quality tables
\usepackage{amsfonts}       % blackboard math symbols
\usepackage{nicefrac}       % compact symbols for 1/2, etc.
\usepackage{microtype}      % microtypography

\usepackage{xcolor}

\usepackage{amsmath}
\usepackage{amssymb}
\usepackage{bm}

\usepackage{float}
\usepackage{graphicx}
\usepackage{caption}
\usepackage{subcaption}

\usepackage{titlesec}      % subsubsections
\usepackage{titlesec,titletoc} 

\usepackage[hang,flushmargin]{footmisc}

\def\<{\langle} \def\>{\rangle}

\usepackage{amsthm}

\newcommand\blfootnote[1]{%
  \begingroup
  \renewcommand\thefootnote{}\footnote{#1}%
  \addtocounter{footnote}{-1}%
  \endgroup
}

\title{Optimal transfer protocol by incremental layer defrosting}

\author{
Federica Gerace$^{1,\star}$, Diego Doimo$^1$, Stefano Sarao Mannelli$^{2}$,\\
Luca Saglietti$^{3}$, and Alessandro Laio$^{1,\star}$
}
\date{
$^1$International School of Advanced Studies (SISSA), Trieste, Italy.\\
$^2$Gatsby Computational Neuroscience Unit \& Sainsbury Wellcome Centre, University College London, London, UK.\\
$^3$Artificial Intelligence Lab, Bocconi University, Milano, Italy \\
}

\usepackage[page]{appendix}

\renewcommand{\appendixtocname}{}

\makeatletter
\let\oldappendix\appendices

\renewcommand{\appendices}{%
  \clearpage
  \renewcommand{\thesection}{\Roman{section}}
  \let\tf@toc\tf@app
  \addtocontents{app}{\protect\setcounter{tocdepth}{5}}
  \immediate\write\@auxout{%
    \string\let\string\tf@toc\string\tf@app^^J
  }
  \oldappendix
}%

\newcommand{\listofappendices}{%
  \begingroup
  \renewcommand{\contentsname}{\appendixtocname}
  \let\@oldstarttoc\@starttoc
  \def\@starttoc##1{\@oldstarttoc{app}}
  \tableofcontents% Reusing the code for \tableofcontents with different \contentsname and different file handle app
  \endgroup
}

\makeatother
\usepackage{imakeidx}
\makeindex
\begin{document}
 \maketitle
 
\blfootnote{$*$To whom correspondence may be addressed. Email: \href{mailto:fgerace@sissa.it}{fgerace@sissa.it}, \href{mailto:alaio@sissa.it}{alaio@sissa.it}.}

%%%%%%%%%%%%%%%%%%%%%%%%%%%%%%%%%%%%%%%%%%%%%%%%%%%%%%%%%

\begin{abstract}
Transfer learning is a powerful tool enabling model training with limited amounts of data. This technique is particularly useful in real-world problems where data availability is often a serious limitation. The simplest transfer learning protocol is based on ``freezing" the feature-extractor layers of a network pre-trained on a data-rich source task, and then adapting only the last layers to a data-poor target task. This workflow is based on the assumption that the feature maps of the pre-trained model are qualitatively similar to the ones that would have been learned with enough data on the target task. 
In this work, we show that this protocol is often sub-optimal, and the largest performance gain may be achieved when smaller portions of the pre-trained network are kept frozen. In particular, we make use of a controlled framework to identify the optimal transfer depth, which turns out to depend non-trivially on the amount of available training data and on the degree of source-target task correlation. We then characterize transfer optimality by analyzing the internal representations of two networks trained from scratch on the source and the target task through multiple established similarity measures. 
\end{abstract}

\section{Introduction}
\label{Introduction}

Machine learning models show a remarkable capacity to extrapolate rules and predict the behavior of complex systems, but this ability often comes at the cost of training with large amounts of data.
In a variety of domains -- such as in medical applications -- collecting data is a slow and costly process or the variability across data is particularly large that we would need huge datasets to achieve satisfying generalization performance \cite{beam2018big,rajkomar2019machine}. This makes data efficiency a necessary condition.
Transfer learning emerged as a solution to this problem \cite{thrun2012learning,shin2016deep,raghu2019transfusion}. By combining a data-poor \textit{target task} with a data-rich related \textit{source task}, it is possible to learn useful representations from the source, and then adapt them to the target in a fine-tuning phase, effectively improving the downstream performance. 
Indeed, fine-tuning the already learned meaningful representation to the target task allows one to work with dramatically smaller dataset sizes while keeping the same level of accuracy. 

Unfortunately, deep neural networks tend to be sensitive even to tiny distribution shifts \cite{gama2014survey}. Therefore, performing transfer learning starting from the right initialization and with the right protocol is crucial for success. 

One of the major risks is to incur in overfitting \cite{geirhos2020shortcut} due to the limited size of the target dataset. Common strategies to avoid it include damping the learning rate in the fine-tuning stage \cite{bengio2012deep}, which forces the network to change its weights only by a small amount. Another popular and simple transfer learning strategy is keeping frozen most of the network \cite{yosinski2014transferable}, and training to the target task only some of the model parameters.
Transferring the layers directly from another network allows, in principle, filtering out irrelevant information present in the input of both the source and the target data sets, effectively reducing the dimensionality of the problem and increasing data efficiency. 

But \emph{how should one choose the layers that should be kept frozen?} Ideally, one should be able to identify, in the source network, a layer in which the representation is general enough to be meaningful also for the target task, but not too general, as otherwise training it to a specialized task might be hindered by the scarcity of data in the target set.  

\begin{figure*}[t!]
\centering
\includegraphics[width=180mm,trim = 0.0cm 6cm 0.0cm 2cm, clip]{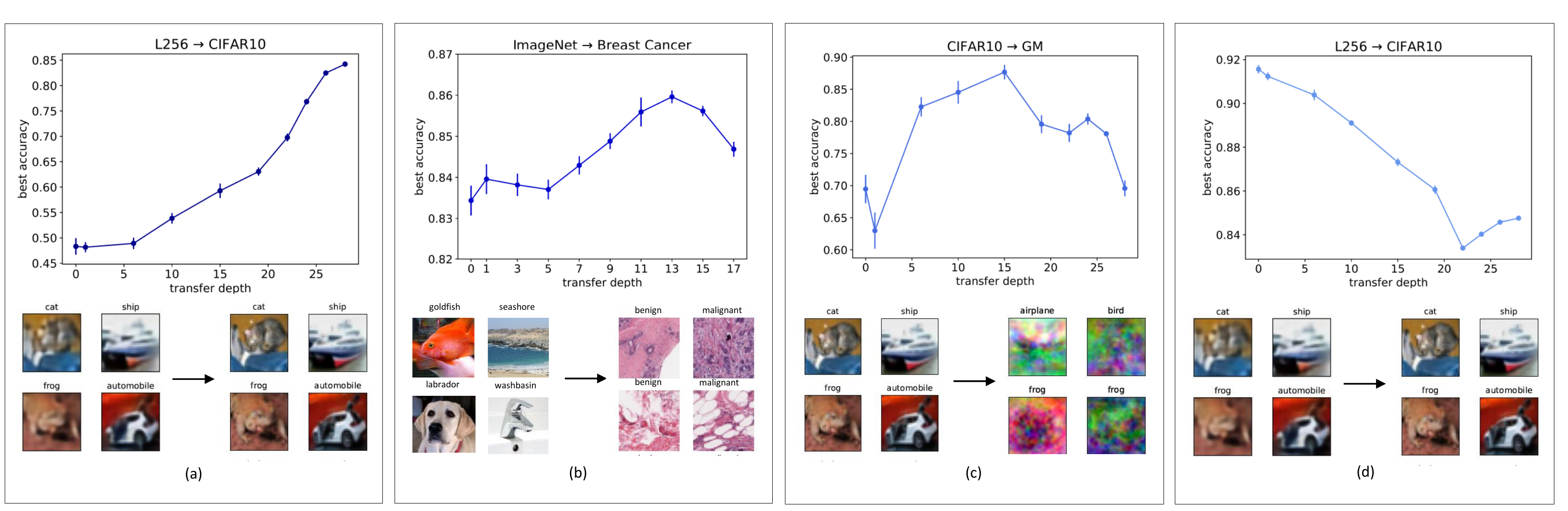}\vspace{-0.5mm}
  \caption{\textbf{Diverse effectiveness of freezing protocols} The figure shows four instances of transfer learning tasks: from  L256 \cite{cifar10} to CIFAR-10 (a,d) with $128$ images per class and full dataset respectively, from ImageNet \cite{deng2009imagenet} to Breast Cancer \cite{breast_cancer_1,breast_cancer_2} with $10^4$ images per class (b), from CIFAR-10 to GM with $32$ images per class(c). We can see samples of each dataset in the lower part of the panels, while the curves above -- the \textit{defrosting profiles} -- represent the test performance of the network. Each point of the curve is the accuracy reached when the first $k$ layers have been transferred. The simulations are averaged over $5$ and $20$ different realizations for the experiments with the synthetic datasets and with the medical datasets respectively.
  }
\label{fig:intro}
\end{figure*}

Given the relevance of transfer learning in real-world applications, it is important to define simple protocols that can be applied without too much fine-tuning or domain knowledge. 
In this work, we propose an approach capable of identifying the layer in which the representation is optimal for a given transfer task. In particular, we suggest estimating the generalization accuracy achieved on the target task as more and more layers are retrained on the target task, while the remaining layers are kept frozen to the feature maps extracted on the source task. 
This procedure allows building what we call a \textit{layer-wise defrosting profile}. Four examples of such profiles are shown in Fig.~\ref{fig:intro}. 
We see that, depending on the transfer scenario, the optimal accuracy can be achieved by either keeping almost all the layers frozen, or by re-optimizing all the layers, or by freezing only part of the layers (panels (a), (d) and (b-c), respectively). 

As we will see, the position of the maximum of the defrosting profiles strongly depends on the nature of the transfer task and on the amount of available data (cf. (a) and (d)). 

Remarkably, in some scenarios, employing a non-vanilla freezing protocol can lead to greater performance improvement than acquiring $5$ times more training data for the target application.

The main results discussed in this work are the following:
 \begin{itemize}
    \vspace{-.3cm}
    \item We introduce a simple protocol capable of identifying, for a given transfer task, the intermediate layer in which the neural network representation is most suitable for transfer; 
    \vspace{-.1cm}
    \item Thanks to a controlled synthetic-dataset framework, we analyze the effect on transfer efficiency of the size of the target task and of the similarity between the source and the target task.
    \vspace{-.1cm}
    \item We show that the large degeneracy of neural network parameter space may obfuscate the similarity and compatibility between two different tasks, and that learning in compliance with both may not hinder performance;
    \vspace{-.1cm}
    \item We show that by monitoring the 
    topological 
    similarity between the representations in networks trained on different tasks it is possible to predict qualitatively if a transfer protocol between the tasks will work or not.
 \end{itemize}

\subsection{Related Works}

Deep neural networks (DNNs) can build surprisingly general representations in a variety of ways \cite{deng2009imagenet, solorio2020review, jaiswal2020survey}. A possible explanation for their effectiveness comes from their "implicit bias" towards regularized solutions \cite{neyshabur2014search}. Numerous works provided evidence of a preference for \textit{simple solutions} \cite{rahaman2018spectral, valle2018deep, shah2020pitfalls, abbe2021staircase}, and it was shown that learning complex functions may require training the networks for longer times \cite{goldt2022datadriven,refinetti2022neural}.

The layered nature of DNN architectures seems to play a crucial role, as it appears that increasingly complex features are progressively encoded in deeper layers of the network \cite{kornblith2019CKA, abbe2021staircase}. Interestingly, the interaction between input data and labels has been shown to lead to non-trivial and often abrupt changes in the nature of the extracted features throughout the layers \cite{tishby2015deep,doimo2020hierarchical}. 
Recent investigations highlighted the importance of intermediate representations \cite{yosinski2014transferable}, but the question on how to select transferable DNN representations \cite{raina2007self,long2015fully} remains largely open, given the complex dependence on the specific source-target distribution shifts and on the data scarcity \cite{lee2022surgical}.  

The theoretical insight on transfer learning phenomenology in DNNs remains limited. Except for some results on training convergence rates \cite{du2020few}, the simplifying assumption of linear activation function was necessary to obtain exact results on the generalization properties after a transfer, both in the asymptotic \cite{dhifallah2021phase, dar2022double} and in the dynamical \cite{lampinen2018analytic} regimes.
A framework for a systematic exploration of the generalization gain as a function of the source-target distribution shift in the non-linear case was developed in \cite{gerace2022probing}, but the analysis was restricted to 2-layer networks. 

The present work explores the uncharted area between theory and real-world transfer learning applications through a carefully designed framework that allows some degree of control over the data distributions. The main goal is to investigate the connection between data and architecture, pinpointing the connection between distribution shift and sample size on one side, and the quality of intermediate representations for a transfer task on the other. 
% %%%%%%%%%%%%%%%%%%%%%%%%%%%%%%%%%%%%%%%%%%%%%%%%%%%%%%%%%%%%%%%%%%

\section{The Experimental setup}\label{sec:exp_setup} 
In this section, we briefly describe the setup used for the experiments of the present paper. We refer to Appendix \ref{app:numerical_details} for more details.

\paragraph*{Architecture and training protocol} In order to assess the robustness of the defrosting profile properties with respect to the network topology, we consider two different types of architectures. The first and mostly employed in our experiments is a Wide-ResNet-28-4 architecture \cite{zagoruyko2016wide}, organized in $3$ identical groups, each containing $4$ blocks of alternating convolutional, batch-norm, and downsampling layers, plus a final batch norm and fully connected one. The reason behind this choice is to foster a good degree of compatibility between model representations, motivated by previous studies \cite{kornblith2019CKA}, where the similarity of neural network representations was shown to increase with the width of the layers. 
We train the Wide-ResNet-28-4 networks for 200 epochs with
stochastic gradient descent with momentum at 0.9, batch size of $128$,
weight decay at $5\cdot10^{-4}$, and a cosine annealing scheduler for the learning rate, starting from a value of $0.1$. 

The second one is a ResNet-18 architecture from the PyTorch model zoo pre-trained on the ImageNet dataset \cite{he2016deep}.  

\paragraph*{Reproducibility}
We provide the code to reproduce our experiments and analysis at \url{https://anonymous.4open.science/r/TL_gradient-DD27/}.

\paragraph*{Synthetic datasets}\label{sec:setup_dataset}
Fig.~\ref{fig:intro} illustrates that defrosting profiles can show different behaviors depending on the specific transfer learning scenario. In order to perform an in-depth investigation of the factors at play in determining these different transfer learning behaviors, we consider three different types of CIFAR-10 clones. These clones are meant to form a hierarchical family of datasets, approximating the true underlying distribution of CIFAR-10 with increasing fidelity.

Following \cite{refinetti2022neural}, the first level of the hierarchy (IsoGM) is obtained by fitting CIFAR-10 with a mixture of isotropic Gaussians and matching the first moments with the CIFAR-10 distribution. At the second level of the hierarchy (GM), also the covariances are learned from CIFAR-10. To go beyond the second moment, we propose to obtain finer approximations of CIFAR-10 images by using a deep autoencoder architecture, consisting of a convolutional encoder network and a deconvolutional decoder, both connected to a bottleneck layer of variable size (we refer to SI for additional details). Note that, by construction, this setup is devoid of the confounding effect of misalignments between labeling rules \cite{gerace2022probing,lee2022surgical} in the source and target tasks, since the synthetic datasets share the same label structure, independently of the alterations induced in the input distributions.

\section{Results}
Basic transfer protocols often prove to be the most effective in real-world applications, since they require very little fine-tuning and allow for straightforward interpretation and comparison of the results. One of the simplest ideas is that of freezing protocols, i.e. constraining some of the layers of the network to the exact values obtained at pre-training and optimizing the remaining layers on the target task. However, even when the analysis is restricted to the family of freezing strategies, in principle one still has to deal with a combinatorial number of possible layer choices \cite{yosinski2014transferable,lee2022surgical}, which makes a simple comparative approach impractical for selecting the best subset of layers to be kept frozen.\\

In this work, we aim to identify the optimal transfer depth, i.e. the layer in which the representation is optimal for a given transfer task. We thus focus on the following type of freezing protocol: we assume the architecture of the network to be left unaltered, with the exception of the last layer, which can be chosen according to the target task (which can be a classification with a different number of categories, or a regression). In the protocol all the first $l$ layers are kept frozen, while the final layers of the network are retrained on the new target data. We call this transfer learning strategy as layer-wise defrosting. One of the key advantages of this protocol is its simplicity. Indeed, no learning heuristics need to be employed to ensure the retention of useful information obtained in the pre-training phase. It also drastically reduces the computational cost of training by pre-processing the dataset a single time, up until the last frozen layer, and then just training the remaining part of the architecture as a separate network receiving the frozen representations as inputs.

In the next three paragraphs, we first describe more in details the layer-wise defrosting strategy, discussing about its effectiveness in the detection of the optimal number of layers to keep frozen. We then show how the optimal transfer depth depends on both the size of the target training set and the degree of similarity between the source and target data statistics.

\begin{figure}[h!]
\hspace{-8mm}
\includegraphics[width=100mm, trim = 8.4cm 0.1cm 7.5cm 0.1cm, clip]{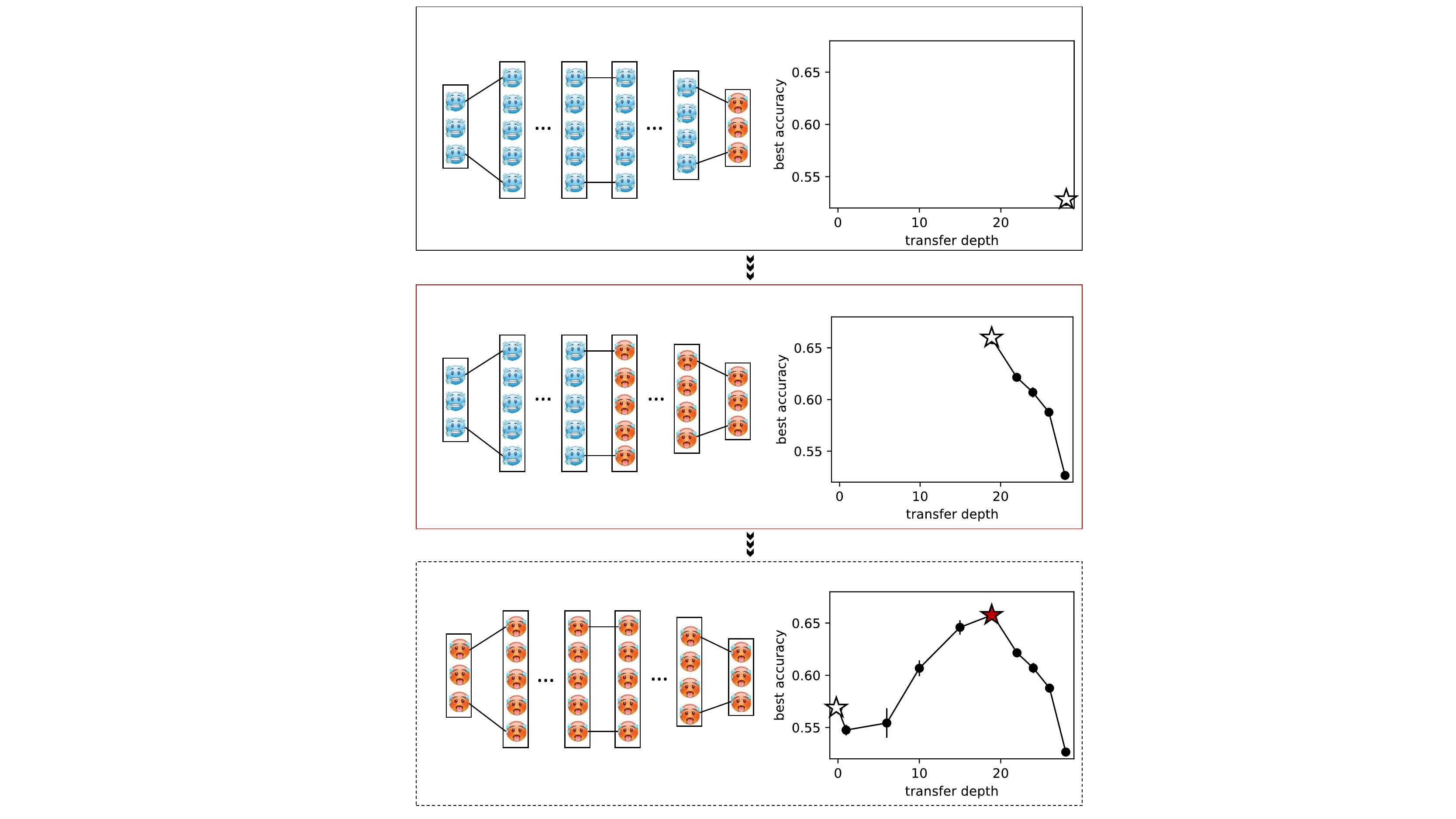}
  \caption{
  \textbf{Layer-wise defrosting procedure.} 
  The cartoon illustrates the transfer procedure employed in the experiments, where the first layers are transferred from the architecture trained on the source and only the last layers are fine-tuned to the target set. On the right, with a star, we show the accuracy reached on the target set by a network trained up to the $k$th layer. 
  }
\label{fig:defrosting}
\end{figure}

\paragraph*{Layer-wise defrosting}

As sketched in the left side of the panels in Fig. ~\ref{fig:defrosting}, given a network architecture and a source/target couple, we perform a series of transfer learning cycles. We start from a completely frozen setting, except the output layer. We then gradually increase the number of layers to be defrosted  and re-trained, starting from the very last layer up to the early layers in the network. At each iteration of this procedure, in order to avoid vanishing gradient scenarios, the defrost layers are re-initialized to random values (following typical heuristics \cite{he2015delving}) and then trained from scratch on the target training data with the hyper-parameter setting described in the Architecture and training protocol paragraph. In the right side of the panels in Fig. ~\ref{fig:defrosting}, we show the accuracy recorded at each iteration of the procedure and, on the bottom, the full \textit{defrosting profile} with the optimal transfer depth highlighted by a red star. 

While it is common practice to transfer most of the network except for the fully-connected segment (or even just the readout layer), in many scenarios transferring a smaller number of layers may largely improve the downstream accuracy. 

There are at least two distinct reasons for the deterioration of the performance when a sub-optimal number of layers is kept frozen. On the one hand, one can see a defrosting profile that decreases when too many layers are kept frozen. This is natural when the input/output mappings associated with the two tasks are too dissimilar: a trained network will not retain all the information contained in the data points, as it learns to select the features  which are meaningful for classification. Some information that is relevant to the target task might be dropped completely by the pre-trained model, given the presence of non-linearities and down-sampling operations throughout the layers. On the other hand, when the target dataset is small, one can see a performance deterioration if too many layers are defrosted and re-trained on the new data. If the ratio of training samples and tunable parameters becomes too small, one can trace over-fitting behaviors, as the model is unable to choose a good generalizing solution among the vast number of zero training error models.

\paragraph*{Dependency of the optimal transfer depth on the training set size}\label{sec:dataset_size}

At a qualitative level the interplay between data scarcity and transfer learning seems clear: with enough samples in the target dataset, all the relevant features of data can be extracted directly from the task at hand. With a reduced sample size, learning a good representations from scratch becomes impossible and transfer learning becomes convenient.

In this section, we quantitatively investigate the transition between these two scenarios, by considering two distinct datasets as test case. The first dataset is the GM CIFAR-10 clone, already described in sec. \ref{sec:exp_setup}.  
The second one is a balanced sub-sample of a publicly available medical dataset for the classification of breast cancer \cite{breast_cancer_1,breast_cancer_2} (we refer to the SI for additional details concerning the Breast Cancer dataset).

Given the two datasets, we then consider two different transfer learning experiments. 
In the first one, we train a Wide-ResNet-28-4 by selecting GM as the source task and CIFAR-10 as the target task. In the second experiment, we take instead a ResNet-18 pre-trained on ImageNet from the PyTorch model zoo and we select as target task the Breast Cancer dataset (this is a common choice in medical applications \cite{khan2019novel,chouhan2020novel}). For both transfer learning experiments, we sub-sample the target data, thus spanning over a wide range of data-scarce scenarios. 

\begin{figure}[h!]
\hspace{-2mm}
\includegraphics[width=93mm]{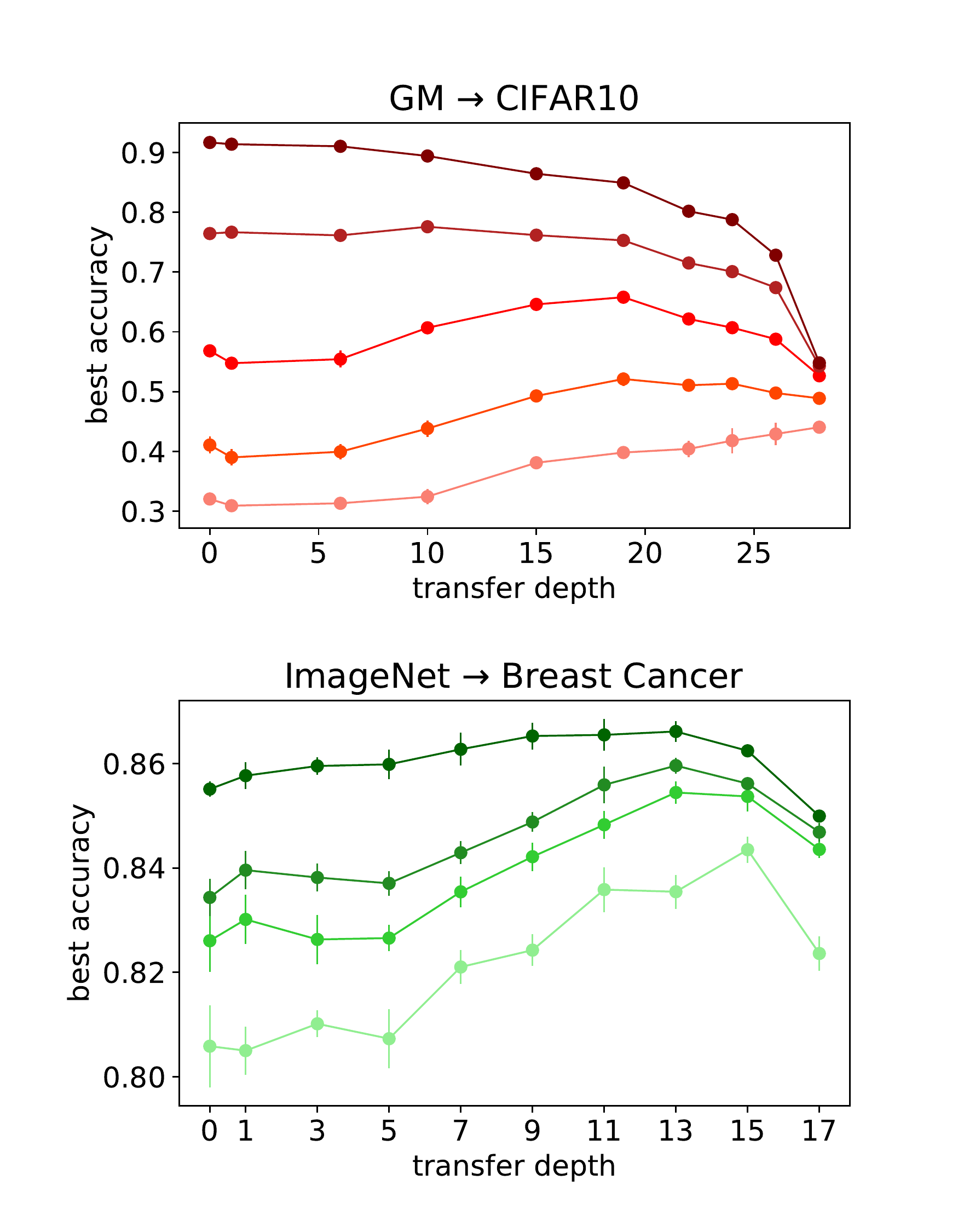}\vspace{-5mm}
  \caption{
  \textbf{Effect of dataset size.}
  The figures show the defrosting curves obtained in two transfer learning tasks for different target set sizes: from GM to CIFAR-10 and from ImageNet to Breast Cancer. The set size is represented by the darkness of the curve, the darker the larger. In particular, in the ImageNet to Breast Cancer transfer, the different shades of green correspond to the training sizes $\{5\times10^2, 5\times10^3, 10^4, 5\times10^4\}$ per class, while, in the GM to CIFAR10 transfer, the different shades of red correspond to the training set sizes $\{2^4, 2^6, 2^8, 2^{10}, 5\times10^3\}$ per class. The simulations are averaged over $20$ and $5$ different realizations in the top and bottom panel respectively. }
\label{fig:dataset_size}
\end{figure}

In Fig.~\ref{fig:dataset_size}, we see the outcome of the experiments involving CIFAR10 (top) and Breast Cancer (bottom) respectively. The defrosting profiles in the two cases show some common features. Indeed, as expected, the generalization accuracy on the target task generally increases with larger amounts of training data. However, a non-trivial observation, is that the optimum of the defrosting profile shifts backward in the layers, as the size of the target training set is increased (darker colors). Instead of just transitioning from a regime of convenient transfers (with some fixed optimal transfer depth) to one of sub-optimal transfer, we find that the number of transferred layers needs to be adjusted according to the sample size, as early layers maintain a higher degree of compatibility when more data is made available. Note also that optimally defrosting the pre-trained network can be more effective than employing the vanilla freezing protocol after acquiring more data. For example, in the bottom panel of Fig.~\ref{fig:dataset_size} the optimal transfer with $5\times10^3$ samples achieves a comparable performance with vanilla freezing with $10$ times more data. 

\paragraph*{Dependency of the optimal transfer depth on source-target similarity} \label{sec:source-target_similarity}

\begin{figure*}[t!]
\hspace{-9mm}
\includegraphics[width=200mm, trim = 0cm 8.5cm 0cm 2.5cm, clip]{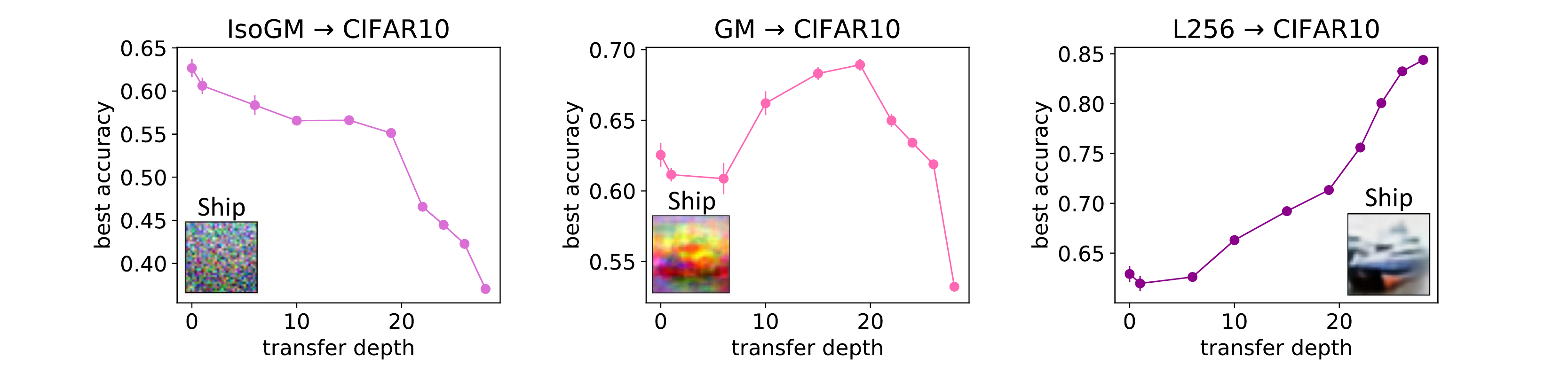}
  \caption{
  \textbf{Defrosting profile for CIFAR-10 clones.}
  The four panels show three different CIFAR-10 clone to CIFAR-10 transfer directions in increasing level of CIFAR-10 approximation: IsoGM, GM, and L256. The figure shows the corresponding defrosting profiles at $384$ training samples per class and provides evidence of how better approximating clones convey richer features and therefore it is best to transfer a larger portion of the architecture. Each of the three insets displays a prototypical examples of how the images looks like in the corresponding clone. The simulations are averaged pver $5$ different realizations.}
\label{fig:clones_accuracy}
\end{figure*}

We now investigate the role played by the statistical similarity between the input distributions of the source and target tasks in determining the optimal transfer depth. Comparing experiments with no control over the input statistics and with different labelling rules, may be insufficient to answer this question, as these factors cannot be easily disentangled. With the hierarchy of generative models described in Section~\ref{sec:setup_dataset}, instead, one can isolate this crucial aspect of transfer learning. In this controlled synthetic framework, we can gradually move from a scenario where only the first moment of the source and target task distribution are matched (IsoGM), to the case in which the first two moments are matched (GM), up to the scenario where the source and target distributions almost perfectly match (L256). 

Our results are summarised in the layer-wise defrosting profiles in Fig.~\ref{fig:clones_accuracy} at same training set size. As expected, if the similarity of the two data sources is insufficient (left panel), fully retraining the network and giving up the transfer learning approach is preferable, even in the data-scarce regime. At pre-training, the early layers of the network  have adapted to significantly different input distributions and are thus unsuitable  to extract the information needed to solve the target task.  
As the similarity between the two tasks is increased (middle and right panel), the optimal cut for the defrosting protocol shifts towards the later layers of the network. In the Appendix, Fig. \ref{fig:other_clones} complements this picture by showing how the defrosting profiles at different size of the training set can significantly change depending on the relatedness between the source and the target task. Indeed, in the case where the source and the target task are poorly statistically correlated, we can observe negative transfer effects even in extreme data-scarcity regimes. On the other hand, when the two tasks are extremely correlated, it is almost always convenient to transfer up to the very last layer, except when one can count on a huge basin of training examples.

From this picture two relevant and related aspects emerge. First of all, these transfer experiments strongly hints at the fact that there is a precise order in which the layers learn different parts of the input distribution. Later layers tend to focus on more complex details and higher-order features of data, selecting only those that are relevant for solving the task at hand. Early layers, as already pointed out\cite{yosinski2014transferable}, seem to play a more generic image pre-processing role, compatible with multiple downstream classification tasks on similar input distributions. The second aspect is that, depending on the complexity of the source task and its relatedness to the target one, we can observe different scenarios, ranging from always negative to almost always positive transfer. It is then natural to ask whether the effectiveness of transfer learning can be assessed by comparing the feature map of two networks trained from scratch on the source and target task respectively. We discuss these two points in the next two sections.

\paragraph*{Compliant Learning}

Given the fact that early layer representations are typically more easily transferable, one wonders if training on different but related image classification tasks leads to networks with similar early layers parameters. As already reported in \cite{kornblith2019CKA}, the answer to this question is typically no: if the network is strongly overparametrized, training on different tasks will lead to different solutions, no matter the task relatedness. However, there are transfer scenarios where the early representations of the network can be completely transferred without hindering the accuracy (cf. the first points of the dark red curves, in the top plot of Fig. \ref{fig:dataset_size}). 
\begin{figure}[h]
\hspace{-2mm}
\includegraphics[width=88mm]{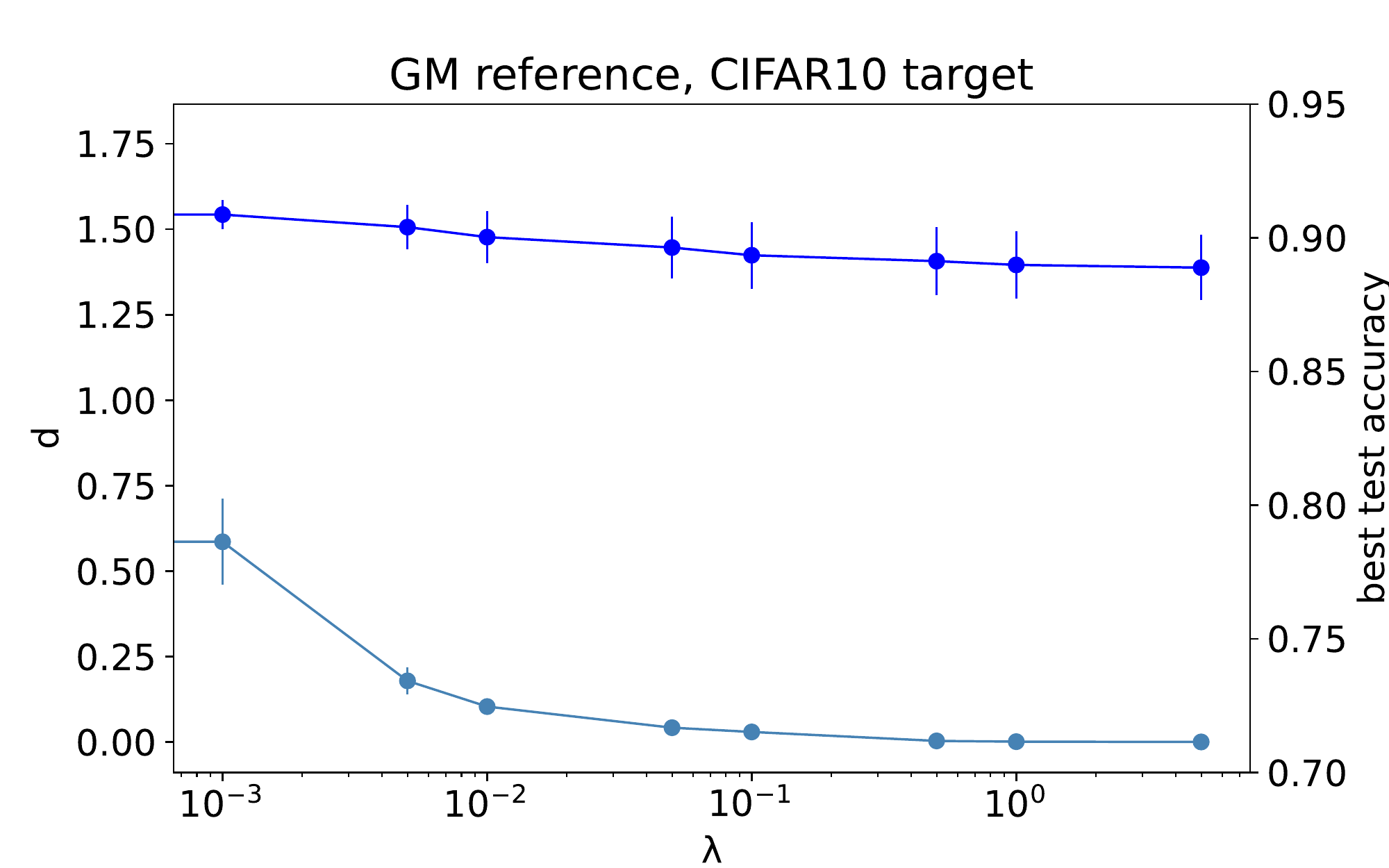}
  \caption{\textbf{Attraction between the network parameters.} The figure shows the generalization accuracy on CIFAR-10 size when the training phase is done in presence of an elastic penalty in the first $10$ layers of the network, centered around the GM pre-trained weights, and the cosine distance between the first layer weights of the source and target network, both trained at full dataset size. 
  %The results show that \emph{compliant learning} is possible in the early layers of the network.
  }
\label{fig:compliant}
\end{figure}

To reconcile these observations, we study the effect of introducing an elastic coupling (i.e. an $L_2$ regularization) of magnitude $\lambda$, which constraints the first $10$ layers of a network trained from scratch on CIFAR10 around the pre-trained values obtained on the GM clone, in the data abundant regime. We then measure the cosine distance between the learned weights of the two networks as a function of the coupling strength $\lambda$.

The outcome is shown in Fig~\ref{fig:compliant}. 
As a preliminary observation, note that when the coupling intensity is set to $0$, the learned weight configurations in the target task are very different from that learned on the source dataset. This can be seen by the very large cosine distance $d\sim0.6$ (light blue curve) in the first layer. Therefore, one could conclude that the process of learning from natural images is not sufficiently strict to constrain the value of the parameters and induce a specific role for the first layers. As expected, by increasing the coupling intensity one can force the weights to converge closer to the reference configuration. What is interesting, however, is that the accuracy (blue curve) is almost  unaffected by the strength of the coupling: the network is able to achieve close to optimal accuracy even when the constraint becomes \emph{hard} and the number of trainable parameters in the second phase is effectively reduced.

We call this procedure ``compliant learning", since, even if the target network is forced to stay close to the source pre-trained weights, this constraint does not downgrade its generalization performance. This procedure is connected to soft-parameter sharing in multi-task learning, with the difference that source and target networks are not trained simultaneously on the respective tasks \cite{caruana1997multitask,ruder2017overview,duong2015low,yang2016deep}.

The arising of compliant learning in this experiment where the source and target datasets share only the first two moments highlights two main aspects. First, it corroborates the finding that the first layer are mostly responsible of learning the lower-order statistics of the data, while the latter layers can focus on higher-order moments, whose relevance is more task-specific. Second, it provides evidence of the great degeneracy of the space of neural network parameters, and of the fact that metrics related to parameter distances may be opaque to meaningful changes in the function represented by the network \cite{kornblith2019CKA}. Indeed, configurations that are apparently very dissimilar can produce equally effective representations. 

In the next section we investigate to what extent the geometry of different representations can be informaccess.
One of the major risks is to incur in overfitting [7]
due to the limited size of the target datasettive of the interchangeability of the neural mappings.

\paragraph*{Representation similarity and transferability}\label{sec:rep_sim} 

Given that metrics based on geometrical arguments in the parameter space may be blind to structural invariance and symmetries of neural network models, if we want to determine the effectiveness of a given transfer direction, we need to move to the feature map space while relying on mostly topological observable.

\emph{How similar are the representations of the target and source network when trained from scratch on the respcetive dataset?} 

To measure this similarity we use the Information Imbalance (II) \cite{Glielmo2022II}. This measure estimates how much the information content of a given feature map is contained into another one. We estimate the II on the test set of the target task. For each data point $i$ in this set we find its nearest neighbor in the representation generated by the network trained on the source task. Say that the nearest neighbor is data point $j$. We then consider the representation of the same data points in the second network (the one trained on the target task). We then find the number of points whose distance from $i$ is smaller than the distance between $i$ and $j$. Denoting this number $n_i$, the II is defined as $\mbox{II}=\frac{2}{N^2}\sum_i n_i$, where $N$ is the number of data in the test set of the target task. If the representations of the two networks are (approximately) equivalent one will find $n_i=0$ for several points and $\mbox{II}\sim 0$. In general, the more the first representation is predictive with respect to the second, the smaller the II is. In the limit in which the first representation does not provide any information on the second, one will find $\mbox{II}=1$ \cite{Glielmo2022II}.
Note that, crucially, the II circumvents all the known ambiguities caused by re-scaling, permutations and symmetries in neural network functions, since this measure relies on the distance ranks, which are invariant.

\begin{figure}[h!]
\hspace{-5mm}
\includegraphics[width=95mm]{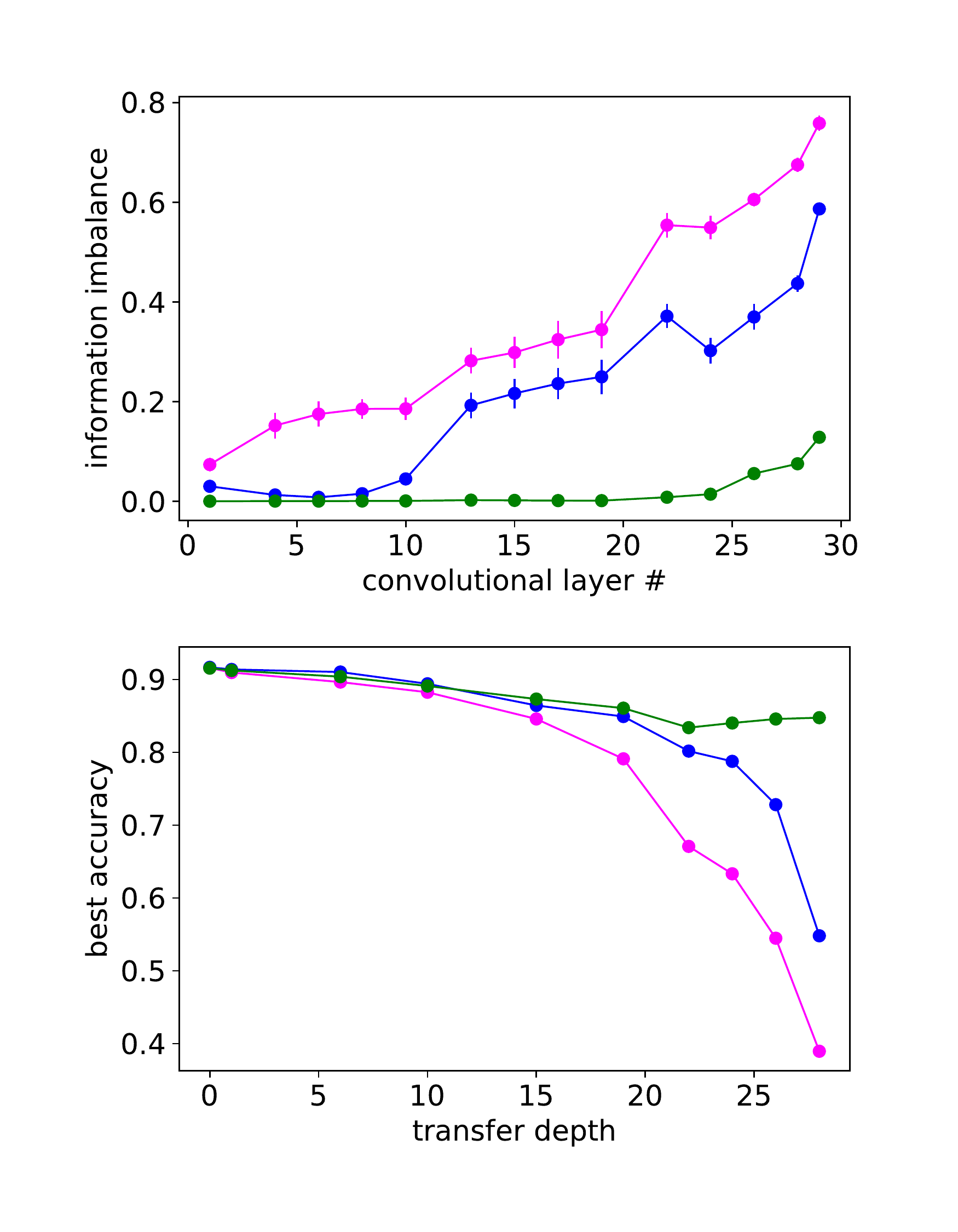} \vspace{-5mm}
  \caption{\textbf{Role of the topological similarity of representations} In the (top) panel we display the Information Imbalance (II) between CIFAR10 and three CIFAR10 clones to be employed as source dataset: IsoGM, GM, and the largest autoencoder with bottleneck $1024$. On the (bottom), we display the layer-wise defrosting profile obtained in these three transfer learning scenarios, assuming source and target datasets are full ($50$K images each). The simulations are averaged over $5$ different realizations. In the bottom panel, the error bars are smaller than point size.
  }
\label{fig:information_imbalance}
\end{figure}

The top panel of Fig.~\ref{fig:information_imbalance} shows the information imbalance quantifying the representation similarity across the layers of two networks trained from scratch on CIFAR10 and on one of its clones respectively, e.g. IsoGM (magenta curve), GM (blue curve) and L256 (green curve). The representation are extracted on the CIFAR10 dataset. In this plot, the first thing to notice is that the II is generally increasing as a function of the depth, but its rate of increase depends on the similarity between the datasets on which the networks to be compared have been trained on. Note that the main II jumps happen in correspondence with the down-sampling layers in the network, where some information that might be relevant for one of the tasks is filtered out in the other task. This inevitably induces different local neighborhoods in the representations. The second thing to notice is the layer at which the representations of CIFAR10 start to consistently deviate from those of its clones. In particular, the topological similarity in the case of L256 is almost perfect until the very last layers, indicating the presence of small differences in the high-order features of the two data distributions. Instead, in the case of GM, we see that the representations are almost equivalent up to the $10$-th layer and become increasingly different thereafter. Finally, the IsoGM case shows that the difference in the statistical distribution of training data affects the learning process dramatically, inducing visibly dissimilar representations that are incompatible with a good generalization performance. The bottom panel of Fig. \ref{fig:information_imbalance} shows instead the defrosting profile obtained when considering CIFAR10 as the target task and one of its clones as the source task (same color legend of the top panel). As we can see, the generalization performance are in agreement with the prediction of the II.  

The behavior of the II supports the understanding that the first layers are responsible for general-purpose pre-processing and are only slightly affected by changes in higher-order statistics of the data distribution. This is compatible with a strong layer-wise hierarchy in the representations extracted by neural networks. This finding complements the picture presented in \cite{goldt2022datadriven}, where the learning dynamics is shown to learn the different moments of the input distribution in a precise temporal order.   

There are many alternative similarity indices that can be measured in place of the II, each one clearly unveiling the connection between the topological similarity of the representations and the transfer learning behavior observed in the layer-wise-defrosting profile. In Appendix \ref{app:similarity_measures} we also show the results for the CKA \cite{kornblith2019CKA}, the neighborhood overlap \cite{doimo2020hierarchical} and the Spearman correlation of local neighborhoods, which provide qualitatively similar information.

\paragraph*{Efficient Defrosting Protocol}

The vanilla freezing protocol, where only the read-out layers are retrained on the target data, may represent a very cost-effective and easily implementable strategy for non-expert practitioners. However, we have shown that a small variation of this protocol, i.e. considering different depths in the defrosting process, may allow large performance gains in the downstream task. 

Given that transfer learning is mostly relevant when the target dataset is effectively small, the multiple defrosting and re-training process should not require high computational costs compared with typical large-scale training of deep networks, and should be compatible with a moderately small computational budget. On the other hand, our results show that the shape of the layer-wise defrosting profile is generally quite regular and predictable from a few points in the profile. Moreover, we have observed that the main drops in the representation similarities happen after the lossy down-sampling layers, while the performance is generally more stable by considering an additional layer of a different type. These findings suggest that the potential gain of defrosting at the correct layer could be uncovered by sampling very few depths and by inferring the positions of the maximum from them. 
In case of a very limited computational budget, we thus suggest sampling a few of these cuts at depths ranging from next to last to after the first down-sampling operation. By looking at the behavior of these points one can obtain an informed guess of what cut is best chosen or what range of layers should include the optimal cut. 

\section{Discussion}

In the present work, we have investigated the relative effectiveness of a family of basic transfer learning protocols. These protocols only differ by the \emph{transfer depth}, i.e. the number of layers which are kept frozen to pre-trained values, while fully optimizing the final layers on the target task. We have shown that the optimal transfer depth depends non-trivially on the amount of training data and on the similarity between source and target tasks, and that its identification can often be more impactful on downstream performance than acquiring additional training data. Crucially, we have also shown that \emph{this depth can be found explicitly by a simple preliminary analysis, } which we dubbed layer-wise defrosting. 
For data-poor target tasks this analysis can be performed at a negligible computational cost, and we propose a strategy that can be employed to reduce its cost also for larger target data sets. We believe that layer-wise defrosting can make a difference in real-world transfer learning applications, as it allows squeezing the best accuracy for a given architecture in an elementary and intuitive manner.

At this time, this type of approach cannot be based on strong theoretical foundations. 

Recent developments in the theory of neural networks \cite{jacot2018neural, canatar2021spectral} achieved exact descriptions of learning neural networks in different regimes, but there is still a lack of mathematical tools for analysing the role played by representations in the feature-learning regime relevant for transfer learning. Empirically, we have shown evidence that the topological similarity between representations -- obtained via independent training on source and target tasks -- provides qualitative indications of the suitability of a representation transfer between tasks. However, we were not able to identify a practical rule of thumb for identifying the best transfer protocol without explicit cross-validation on a target test set.

In perspective, we plan to develop a training schedule for transfer learning capable of efficiently performing layer-wise defrosting in a fully automatic manner. This could be achieved, for example, by exploiting the property which we called compliant learning, namely the fact that transferable representations are practically insensitive to $L_2$ penalties between the weights. More in general, we believe that understanding -- both on the theoretical and the practical sides -- how to disentangle input similarity from the labeling rule might be an interesting direction for future investigations in this field of research.

\section*{Acknowledgements}
The authors acknowledge Sebastian Goldt for useful discussions. SSM acknowledges support from the Sainsbury Wellcome Centre Core Grant from Wellcome (219627/Z/19/Z) and the Gatsby Charitable Foundation (GAT3755).

\bibliography{ref}

\begin{thebibliography}{10}

\bibitem{beam2018big}
Andrew~L Beam and Isaac~S Kohane.
\newblock Big data and machine learning in health care.
\newblock {\em Jama}, 319(13):1317--1318, 2018.

\bibitem{rajkomar2019machine}
Alvin Rajkomar, Jeffrey Dean, and Isaac Kohane.
\newblock Machine learning in medicine.
\newblock {\em New England Journal of Medicine}, 380(14):1347--1358, 2019.

\bibitem{thrun2012learning}
Sebastian Thrun and Lorien Pratt.
\newblock {\em Learning to learn}.
\newblock Springer Science \& Business Media, 2012.

\bibitem{shin2016deep}
Hoo-Chang Shin, Holger~R Roth, Mingchen Gao, Le~Lu, Ziyue Xu, Isabella Nogues,
  Jianhua Yao, Daniel Mollura, and Ronald~M Summers.
\newblock Deep convolutional neural networks for computer-aided detection: Cnn
  architectures, dataset characteristics and transfer learning.
\newblock {\em IEEE transactions on medical imaging}, 35(5):1285--1298, 2016.

\bibitem{raghu2019transfusion}
Maithra Raghu, Chiyuan Zhang, Jon Kleinberg, and Samy Bengio.
\newblock Transfusion: Understanding transfer learning for medical imaging.
\newblock {\em Advances in neural information processing systems}, 32, 2019.

\bibitem{gama2014survey}
Jo{\~a}o Gama, Indr{\.e} {\v{Z}}liobait{\.e}, Albert Bifet, Mykola Pechenizkiy,
  and Abdelhamid Bouchachia.
\newblock A survey on concept drift adaptation.
\newblock {\em ACM computing surveys (CSUR)}, 46(4):1--37, 2014.

\bibitem{geirhos2020shortcut}
Robert Geirhos, J{\"o}rn-Henrik Jacobsen, Claudio Michaelis, Richard Zemel,
  Wieland Brendel, Matthias Bethge, and Felix~A Wichmann.
\newblock Shortcut learning in deep neural networks.
\newblock {\em Nature Machine Intelligence}, 2(11):665--673, 2020.

\bibitem{bengio2012deep}
Yoshua Bengio.
\newblock Deep learning of representations for unsupervised and transfer
  learning.
\newblock In {\em Proceedings of ICML workshop on unsupervised and transfer
  learning}, pages 17--36. JMLR Workshop and Conference Proceedings, 2012.

\bibitem{yosinski2014transferable}
Jason Yosinski, Jeff Clune, Yoshua Bengio, and Hod Lipson.
\newblock How transferable are features in deep neural networks?
\newblock {\em Advances in neural information processing systems}, 27, 2014.

\bibitem{cifar10}
Alex Krizhevsky, Vinod Nair, and Geoffrey Hinton.
\newblock Cifar-10 (canadian institute for advanced research).

\bibitem{deng2009imagenet}
Jia Deng, Wei Dong, Richard Socher, Li-Jia Li, Kai Li, and Li~Fei-Fei.
\newblock Imagenet: A large-scale hierarchical image database.
\newblock In {\em 2009 IEEE conference on computer vision and pattern
  recognition}, pages 248--255. Ieee, 2009.

\bibitem{breast_cancer_1}
Andrew Janowczyk and Anant Madabhushi.
\newblock Deep learning for digital pathology image analysis: A comprehensive
  tutorial with selected use cases.
\newblock {\em Journal of pathology informatics}, 7(1):29, 2016.

\bibitem{breast_cancer_2}
Angel Cruz-Roa, Ajay Basavanhally, Fabio Gonz{\'a}lez, Hannah Gilmore, Michael
  Feldman, Shridar Ganesan, Natalie Shih, John Tomaszewski, and Anant
  Madabhushi.
\newblock Automatic detection of invasive ductal carcinoma in whole slide
  images with convolutional neural networks.
\newblock In {\em Medical Imaging 2014: Digital Pathology}, volume 9041, page
  904103. SPIE, 2014.

\bibitem{solorio2020review}
Sa{\'u}l Solorio-Fern{\'a}ndez, J~Ariel Carrasco-Ochoa, and Jos{\'e}~Fco
  Mart{\'\i}nez-Trinidad.
\newblock A review of unsupervised feature selection methods.
\newblock {\em Artificial Intelligence Review}, 53(2):907--948, 2020.

\bibitem{jaiswal2020survey}
Ashish Jaiswal, Ashwin~Ramesh Babu, Mohammad~Zaki Zadeh, Debapriya Banerjee,
  and Fillia Makedon.
\newblock A survey on contrastive self-supervised learning.
\newblock {\em Technologies}, 9(1):2, 2020.

\bibitem{neyshabur2014search}
Behnam Neyshabur, Ryota Tomioka, and Nathan Srebro.
\newblock In search of the real inductive bias: On the role of implicit
  regularization in deep learning.
\newblock {\em arXiv preprint arXiv:1412.6614}, 2014.

\bibitem{rahaman2018spectral}
Nasim Rahaman, Devansh Arpit, Aristide Baratin, Felix Draxler, Min Lin, Fred~A
  Hamprecht, Yoshua Bengio, and Aaron~C Courville.
\newblock On the spectral bias of deep neural networks.
\newblock 2018.

\bibitem{valle2018deep}
Guillermo~Valle P{\'{e}}rez, Chico~Q. Camargo, and Ard~A. Louis.
\newblock Deep learning generalizes because the parameter-function map is
  biased towards simple functions.
\newblock In {\em 7th International Conference on Learning Representations,
  {ICLR} 2019, New Orleans, LA, USA, May 6-9, 2019}. OpenReview.net, 2019.

\bibitem{shah2020pitfalls}
Harshay Shah, Kaustav Tamuly, Aditi Raghunathan, Prateek Jain, and Praneeth
  Netrapalli.
\newblock The pitfalls of simplicity bias in neural networks.
\newblock {\em Advances in Neural Information Processing Systems},
  33:9573--9585, 2020.

\bibitem{abbe2021staircase}
Emmanuel Abbe, Enric Boix-Adsera, Matthew~S Brennan, Guy Bresler, and Dheeraj
  Nagaraj.
\newblock The staircase property: How hierarchical structure can guide deep
  learning.
\newblock {\em Advances in Neural Information Processing Systems},
  34:26989--27002, 2021.

\bibitem{goldt2022datadriven}
Sebastian Goldt and Alessandro Ingrosso.
\newblock Data-driven emergence of convolutional structure in neural networks.
\newblock {\em Proceedings of the National Academy of Sciences},
  119(40):2201854119, 2022.

\bibitem{refinetti2022neural}
Maria Refinetti, Alessandro Ingrosso, and Sebastian Goldt.
\newblock Neural networks trained with sgd learn distributions of increasing
  complexity.
\newblock {\em arXiv preprint arXiv:2211.11567}, 2022.

\bibitem{kornblith2019CKA}
Simon Kornblith, Mohammad Norouzi, Honglak Lee, and Geoffrey Hinton.
\newblock Similarity of neural network representations revisited.
\newblock In {\em International Conference on Machine Learning}, pages
  3519--3529. PMLR, 2019.

\bibitem{tishby2015deep}
Naftali Tishby and Noga Zaslavsky.
\newblock Deep learning and the information bottleneck principle.
\newblock In {\em 2015 ieee information theory workshop (itw)}, pages 1--5.
  IEEE, 2015.

\bibitem{doimo2020hierarchical}
Diego Doimo, Aldo Glielmo, Alessio Ansuini, and Alessandro Laio.
\newblock Hierarchical nucleation in deep neural networks.
\newblock {\em Advances in Neural Information Processing Systems},
  33:7526--7536, 2020.

\bibitem{raina2007self}
Rajat Raina, Alexis Battle, Honglak Lee, Benjamin Packer, and Andrew~Y Ng.
\newblock Self-taught learning: transfer learning from unlabeled data.
\newblock In {\em Proceedings of the 24th international conference on Machine
  learning}, pages 759--766, 2007.

\bibitem{long2015fully}
Jonathan Long, Evan Shelhamer, and Trevor Darrell.
\newblock Fully convolutional networks for semantic segmentation.
\newblock In {\em Proceedings of the IEEE conference on computer vision and
  pattern recognition}, pages 3431--3440, 2015.

\bibitem{lee2022surgical}
Yoonho Lee, Annie~S Chen, Fahim Tajwar, Ananya Kumar, Huaxiu Yao, Percy Liang,
  and Chelsea Finn.
\newblock Surgical fine-tuning improves adaptation to distribution shifts.
\newblock {\em arXiv preprint arXiv:2210.11466}, 2022.

\bibitem{du2020few}
Simon~S Du, Wei Hu, Sham~M Kakade, Jason~D Lee, and Qi~Lei.
\newblock Few-shot learning via learning the representation, provably.
\newblock {\em arXiv preprint arXiv:2002.09434}, 2020.

\bibitem{dhifallah2021phase}
Oussama Dhifallah and Yue~M. Lu.
\newblock Phase transitions in transfer learning for high-dimensional
  perceptrons.
\newblock {\em Entropy}, 23(4):400, 2021.

\bibitem{dar2022double}
Yehuda Dar and Richard~G Baraniuk.
\newblock Double double descent: on generalization errors in transfer learning
  between linear regression tasks.
\newblock {\em SIAM Journal on Mathematics of Data Science}, 4(4):1447--1472,
  2022.

\bibitem{lampinen2018analytic}
Andrew~K. Lampinen and Surya Ganguli.
\newblock An analytic theory of generalization dynamics and transfer learning
  in deep linear networks.
\newblock In {\em 7th International Conference on Learning Representations,
  {ICLR} 2019, New Orleans, LA, USA, May 6-9, 2019}. OpenReview.net, 2019.

\bibitem{gerace2022probing}
Federica Gerace, Luca Saglietti, Stefano~Sarao Mannelli, Andrew Saxe, and Lenka
  Zdeborov{\'a}.
\newblock Probing transfer learning with a model of synthetic correlated
  datasets.
\newblock {\em Machine Learning: Science and Technology}, 3(1):015030, 2022.

\bibitem{zagoruyko2016wide}
S.~Zagoruyko and N.~Komodakis.
\newblock Wide residual networks.
\newblock {\em arXiv preprint arXiv:1605.07146}, 2016.

\bibitem{he2016deep}
Kaiming He, Xiangyu Zhang, Shaoqing Ren, and Jian Sun.
\newblock Deep residual learning for image recognition.
\newblock In {\em Proceedings of the IEEE conference on computer vision and
  pattern recognition}, pages 770--778, 2016.

\bibitem{he2015delving}
Kaiming He, Xiangyu Zhang, Shaoqing Ren, and Jian Sun.
\newblock Delving deep into rectifiers: Surpassing human-level performance on
  imagenet classification.
\newblock In {\em Proceedings of the IEEE international conference on computer
  vision}, pages 1026--1034, 2015.

\bibitem{khan2019novel}
SanaUllah Khan, Naveed Islam, Zahoor Jan, Ikram~Ud Din, and Joel JP~C
  Rodrigues.
\newblock A novel deep learning based framework for the detection and
  classification of breast cancer using transfer learning.
\newblock {\em Pattern Recognition Letters}, 125:1--6, 2019.

\bibitem{chouhan2020novel}
Vikash Chouhan, Sanjay~Kumar Singh, Aditya Khamparia, Deepak Gupta, Prayag
  Tiwari, Catarina Moreira, Robertas Dama{\v{s}}evi{\v{c}}ius, and Victor
  Hugo~C De~Albuquerque.
\newblock A novel transfer learning based approach for pneumonia detection in
  chest x-ray images.
\newblock {\em Applied Sciences}, 10(2):559, 2020.

\bibitem{caruana1997multitask}
Rich Caruana.
\newblock Multitask learning.
\newblock {\em Machine learning}, 28(1):41--75, 1997.

\bibitem{ruder2017overview}
Sebastian Ruder.
\newblock An overview of multi-task learning in deep neural networks.
\newblock {\em arXiv preprint arXiv:1706.05098}, 2017.

\bibitem{duong2015low}
Long Duong, Trevor Cohn, Steven Bird, and Paul Cook.
\newblock Low resource dependency parsing: Cross-lingual parameter sharing in a
  neural network parser.
\newblock In {\em Proceedings of the 53rd annual meeting of the Association for
  Computational Linguistics and the 7th international joint conference on
  natural language processing (volume 2: short papers)}, pages 845--850, 2015.

\bibitem{yang2016deep}
Yongxin Yang and Timothy Hospedales.
\newblock Deep multi-task representation learning: A tensor factorisation
  approach.
\newblock {\em arXiv preprint arXiv:1605.06391}, 2016.

\bibitem{Glielmo2022II}
Aldo Glielmo, Claudio Zeni, Bingqing Cheng, Gábor Csányi, and Alessandro
  Laio.
\newblock {Ranking the information content of distance measures}.
\newblock {\em PNAS Nexus}, 1(2), 04 2022.
\newblock pgac039.

\bibitem{jacot2018neural}
Arthur Jacot, Franck Gabriel, and Cl{\'e}ment Hongler.
\newblock Neural tangent kernel: Convergence and generalization in neural
  networks.
\newblock {\em Advances in neural information processing systems}, 31, 2018.

\bibitem{canatar2021spectral}
Abdulkadir Canatar, Blake Bordelon, and Cengiz Pehlevan.
\newblock Spectral bias and task-model alignment explain generalization in
  kernel regression and infinitely wide neural networks.
\newblock {\em Nature communications}, 12(1):1--12, 2021.

\bibitem{lippe2022uvadlc}
Phillip Lippe.
\newblock {UvA Deep Learning Tutorials}.
\newblock \url{https://uvadlc-notebooks.readthedocs.io/en/latest/}, 2022.

\bibitem{kingma2014adam}
Diederik~P Kingma and Jimmy Ba.
\newblock Adam: A method for stochastic optimization.
\newblock {\em arXiv preprint arXiv:1412.6980}, 2014.

\end{thebibliography}
\bibliographystyle{unsrt}

%%%%%%%%%%%%%%%%%%%%%%%%%%%%%%%%%%%%%%%%%%%%%%%%%%%%%%%%%%%%%%%%%%%%%%%%%%%%%%%
%%%%%%%%%%%%%%%%%%%%%%%%%%%%%%%%%%%%%%%%%%%%%%%%%%%%%%%%%%%%%%%%%%%%%%%%%%%%%%%
% APPENDIX
%%%%%%%%%%%%%%%%%%%%%%%%%%%%%%%%%%%%%%%%%%%%%%%%%%%%%%%%%%%%%%%%%%%%%%%%%%%%%%%
%%%%%%%%%%%%%%%%%%%%%%%%%%%%%%%%%%%%%%%%%%%%%%%%%%%%%%%%%%%%%%%%%%%%%%%%%%%%%%%
\newpage
\appendix
\onecolumn

\section{Numerical Details} \label{app:numerical_details}

In this section we provide extra-details concerning both the autoencoder clones and the Breast Cancer medical dataset.

\paragraph*{Autoencoder Clones}
We follow \cite{lippe2022uvadlc} to build an encoder with 5 $3\times3$-convolutional layers and GeLU activations, alternating stride 2 and stride 1 filters to reduce the dimensionality and ending with a single dense layer, whose size determines the bottleneck. The decoder has the same structure with the stride-2 convolutional layers replaced by deconvolutional ones. Training is performed over the CIFAR-10 training set in batches of size 256 over 500 epochs using Adam \cite{kingma2014adam}. We used a warmup schedule, peaking the learning rate up to $10^{-3}$ after 100 epochs and then annealing to $10^{-5}$. At the end of the training, we collect the produced images for each bottleneck size in a new dataset. In the experiments, the so-produced CIFAR-10 autoencoder clones are standardized in such a way to have zero mean and standard deviation equal to one for each channel.  

\paragraph*{Medical Dataset}
Concerning the transfer experiments in the bottom panel of Fig. \ref{fig:dataset_size}, we consider as test case a publicly available medical dataset for the classification of breast cancer, consisting of 277.524 50x50 patches extracted from 162 scans \cite{breast_cancer_1,breast_cancer_2}. The images are labeled as positive or negative according to the presence or absence of invasive ductal carcinoma (IDC) tissue in the patch. Among all the examples, we use the $80\%$ of them for the training set and the $20\%$ for the test set. With the purpose of speeding-up the simulations and describing more realistic transfer learning scenarios, we further randomly sub-sampled the test set up to $10^4$ images at each simulation run, preserving class balance. In the transfer experiment, the images have been rescaled to $224\times 224$ pixels and then standardized as required in the PyTorch documentation for pre-trained ResNet-18.

\section{More on CIFAR10 Clones: Defrosting Profile and Information Imbalance}

As already pointed out in the main text, the optimal transfer depth strongly depends on the size of the target training set and on the statistical similarity between the source and the target task. 

\begin{figure}[h!]
\centering
\includegraphics[width=150mm]{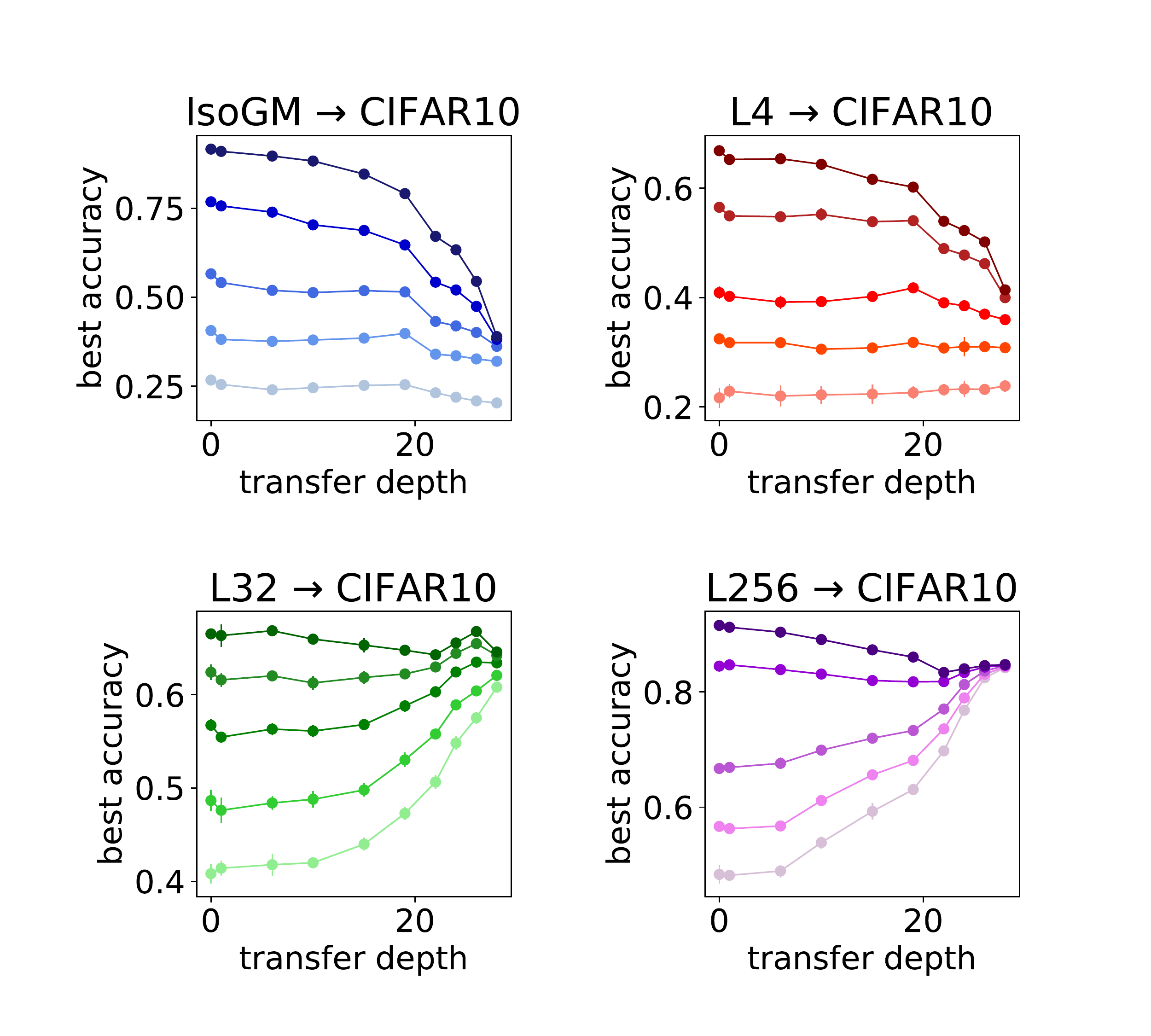}
  \caption{ \textbf{Effect of dataset size and source-target relatedness.} The plot shows the defrosting profiles at varying training set size of CIFAR-10 for four different transfer couples, similarly to what was shown in Fig.~\ref{fig:dataset_size} (top panel) of the main text. The transfers reported are: IsoGM to CIFAR-10 (top left panel), L4 to  CIFAR-10 (top right panel), L32 to CIFAR-10 (bottom left panel) and L256 to CIFAR-10 (bottom right panel). In particular, in the Iso-GM to CIFAR-10 transfer direction, the curves are relative from the lighter to the darker colors to the following CIFAR-10 training set sizes $\{8, 64, 256, 1024, \mbox{full dataset size}\}$; in the L4 to CIFAR-10 transfer direction, the curves are relative from the lighter to the darker colors to the following CIFAR-10 training set sizes $\{4, 16, 64, 256, 512\}$ ; in the L32 to CIFAR-10 transfer direction, the curves are relative from the lighter to the darker colors to the following CIFAR-10 training set sizes $\{64, 128, 256, 384, 512\}$; in the L256 to CIFAR-10 transfer direction, the curves are relative from the lighter to the darker colors to the following CIFAR-10 training set sizes $\{128, 256, 512, 2048, \mbox{full dataset size}\}$. The simulations are averaged over $5$ different realizations and the error bars are smaller than the point size.}
\label{fig:other_clones}
\end{figure}

Fig. \ref{fig:other_clones} complements the picture described in sec.\ref{sec:dataset_size} and sec.\ref{sec:source-target_similarity}, by showing the defrosting profile for different CIFAR-10 training set size (lighter colors identify smaller training set size), when the source task is IsoGM or one of the autoencoder clones at increasingly higher bottleneck size, e.g. L4, L32, L256. As we can see, in the transfer direction IsoGM to CIFAR-10, transferring up to the very layer is never beneficial, no matter the size of the target task. IsoGM and CIFAR-10 are indeed sharing only the first moment of the underlying CIFAR-10 distribution, therefore, negative transfer effects can occur due to the scarce statistical similarity between the two datasets. By including higher order moments of the CIFAR-10 distribution, and refining their approximations through the bottleneck size of the CIFAR-10 autoencoder clones, transfer learning becomes increasingly more effective. Indeed, in the transfer direction L256 to CIFAR-10, we can see it is always convenient to transfer up to the very last layer, except at very huge training set sizes. Almost the same picture occurs when transferring the pre-trained feature map on L32 to CIFAR10. However, similarly to the picture emerged in the transfer experiment concerning the Breast Cancer dataset, in this case, a peak starts appearing in correspondence of the second-to-last defrosted layer. The optimal transfer depth then shifts to the left in the L4 to CIFAR-10 transfer experiment with $64$ images per class. This phenomenon is probably due to the smaller degree of similarity between the two datasets, even if it is less exacerbated then what it is observable in the GM to CIFAR-10 transfer direction. 
 
\begin{figure}[h!]
\centering
\includegraphics[width=100mm]{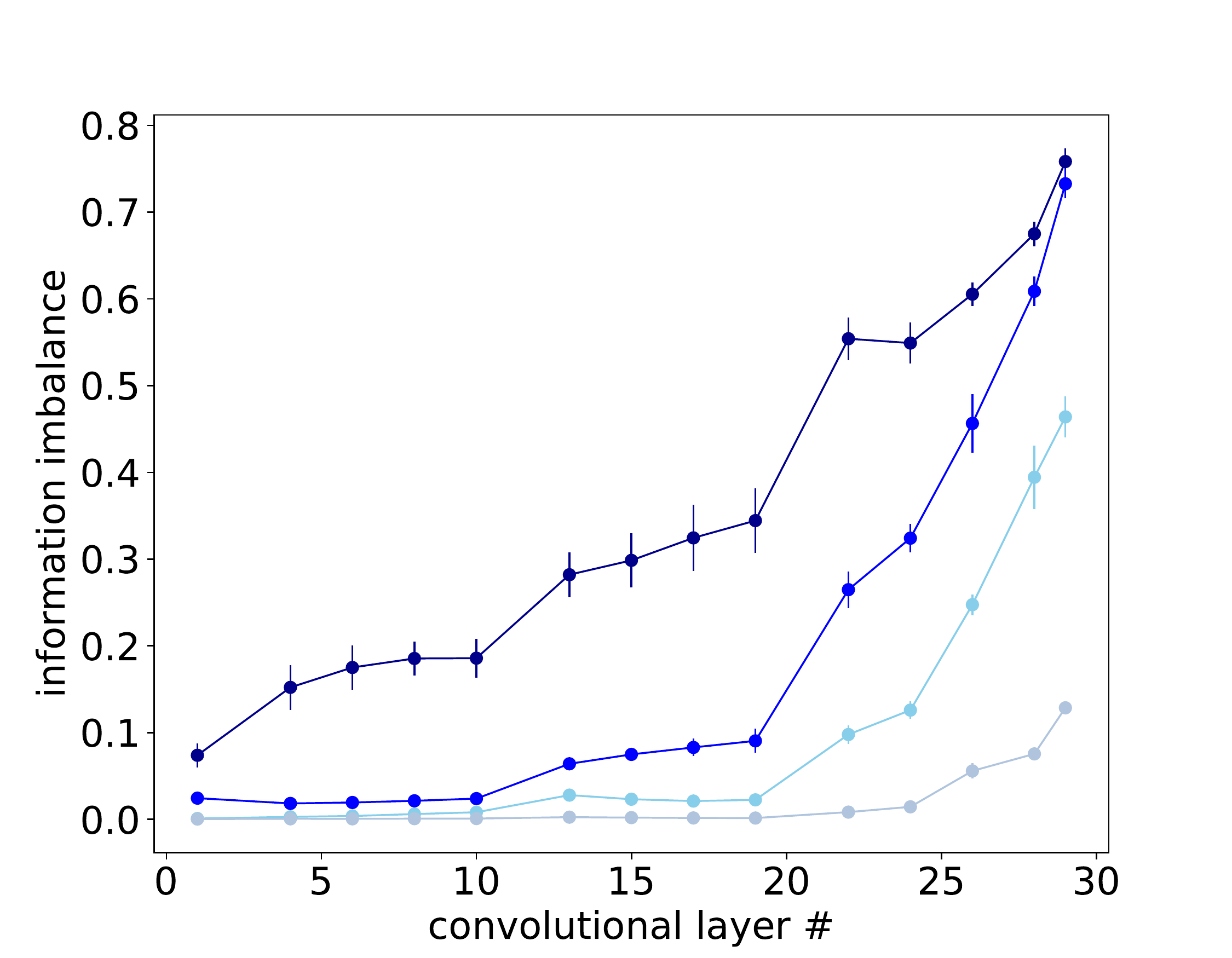}
  \caption{\textbf{Representation Similarity.} The plot displays the information imbalance across the layers for increasing fidelity of the CIFAR-10 clones. In particular, lighter colors refer to more similar clones, (in sequence IsoGM, L4, L32, L256). The source and target network are trained at full dataset size (50K images). The simulations are averaged over $5$ different realizations.}
\label{fig:ii_autoclones}
\end{figure}

Fig. \ref{fig:ii_autoclones} shows the II across layers, quantifying the representation similarity of two networks trained from scratch on CIFAR-10 and on one of the clones among IsoGM, L4, L32 and L256. The representations are extracted on the CIFAR-10 test set. As we can see, the profile of the II reflect the hierarchical nature of the CIFAR-10 clone family. Indeed, the larger is the size of the bottleneck, the more and the deeper similar the representations are across the layers.

\section{Similarity measures} \label{app:similarity_measures}

In the main manuscript, we have analyzed the similarity among representations while relying on the II metrics. However, as already pointed out in sec.\ref{sec:rep_sim}, there are other similarity measures that can be taken into account and that are equivalently blind to some of the re-scaling, permutation and symmetries of the parametric function implemented by a neural network model.

\begin{figure}[h!]
\includegraphics[width=\textwidth]{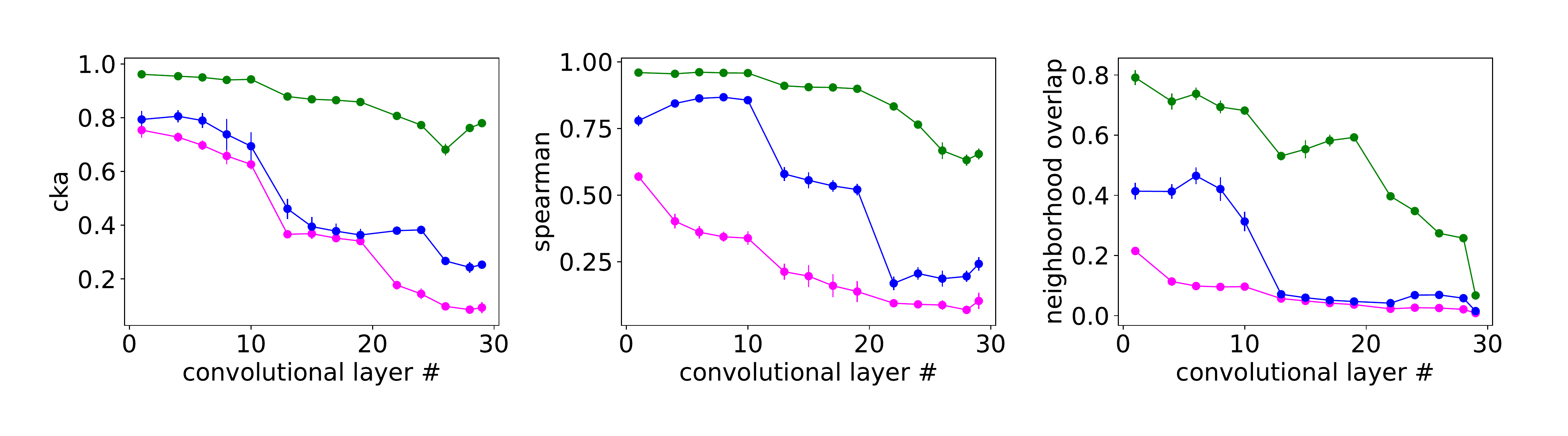}
  \caption{\textbf{Other Similarity Measures.} From left to right, CKA, Spearman Correlation and Neighborhood Overlap across layers. The different colors refer to different source tasks, e.g. IsoGM (magenta curve), GM (blue curve) and L256 (green curve). The setting is the same as reported in the main text Fig.~\ref{fig:information_imbalance}, i.e. source and target networks are trained on the full training set size (50K images). The simulations are averaged over $5$ different realizations.}
\label{fig:similarity_measures}
\end{figure}

Fig. \ref{fig:similarity_measures} shows, from left to right, the CKA \cite{kornblith2019CKA}, the Spearman correlation of the first $100$ neighborhoods and the neighborhood overlap among the first $30$ \cite{doimo2020hierarchical} neighbours as a function of the network depth. The representations are extracted once again on the CIFAR-10 test set and corresponds to the feature map of two different networks, one trained from scratch on the CIFAR-10 dataset and the other one on one of the CIFAR-10 clones, e.g. IsoGM (magenta curve), GM (blue curve) and L256 (green curve). As can be seen, the behaviour of the three metrics is qualitatively similar to the one already observed for the II in Fig. \ref{fig:information_imbalance} (top panel). Indeed, even in this case, the representation similarity is stronger at early layers, providing a further hint on the generality of the feature maps extracted at the bottom of the networks. Moreover, the hierarchical structure of the CIFAR10-clones is reflected in the profiles of the similarity measured, signaling which task is the most promising one as source task for the CIFAR-10 classification target task.  

%%%%%%%%%%%%%%%%%%%%%%%%%%%%%%%%%%%%%%%%%%%%%%%%%%%%%%%%%%%%

\end{document}